\documentclass[11pt]{article}
\usepackage{emnlp2017}
\usepackage{times}
\usepackage{url}
\usepackage{paralist}
\usepackage{latexsym}
\usepackage{ucs}
\usepackage{fontenc}
\usepackage[utf8x]{inputenc}
\makeatletter
\def\@arabic#1{\number#1}
\makeatother
\usepackage[english]{babel}
\usepackage{dsfont}
\usepackage{textcomp}
\usepackage{amsfonts}
\usepackage{amsmath}
\usepackage{graphicx}
\usepackage{sidecap}
\usepackage{pgfplots}
\pgfplotsset{compat=1.10}
\usepackage{array} 
\usepackage{algorithm}
\usepackage[noend]{algpseudocode}
\usepackage{multirow}
\usepackage{bm}
\usepackage{breqn}
\usepackage{booktabs}

\emnlpfinalcopy

\makeatletter
\newcommand{\@emptybiblabel}[1]{}
\makeatother

\newcommand{\com}[1]{}

\newcommand{\varembed}[0]{VarEmbed} 
\newcommand{\Ouralgo}[0]{\textsc{mimick}}
\newcommand{\ouralgo}[0]{\textsc{mimick}} 

\newcommand{\msd}[0]{attributes} 
\newcommand{\chartotag}[0]{\textsc{char$\to$tag}}
\newcommand{\morsyn}[0]{morphosyntactic}
\newcommand{\Morsyn}[0]{Morphosyntactic}

\newcommand{\none}[0]{\textit{NONE}}
\newcommand{\rnn}{\textsc{rnn}} 

\renewcommand{\vec}[1]{\bm{#1}}
\newcommand{\vh}{\ensuremath \vec{h}}
\newcommand{\vb}{\ensuremath \vec{b}}
\newcommand{\ve}{\ensuremath \vec{e}}

\title{Mimicking Word Embeddings using Subword RNNs}

\author{Yuval Pinter \qquad ~~~~ Robert Guthrie \qquad ~~~~ Jacob Eisenstein \\ \\
  School of Interactive Computing \\
  Georgia Institute of Technology \\
        \{uvp,rguthrie3,jacobe\}@gatech.edu }

\date{\today}

\begin{document}

\maketitle
\begin{abstract}
Word embeddings improve generalization over lexical features by placing each word in a lower-dimensional space, using distributional information obtained from unlabeled data.
However, the effectiveness of word embeddings for downstream NLP tasks is limited by out-of-vocabulary (OOV) words, for which embeddings do not exist.
In this paper, we present \ouralgo, an approach to generating OOV word embeddings compositionally, by learning a function from spellings to distributional embeddings.
Unlike prior work, \ouralgo{} does not require re-training on the original word embedding corpus; instead, learning is performed at the type level.
Intrinsic and extrinsic evaluations demonstrate the power of this simple approach. 
On 23 languages, \ouralgo{} improves performance over a word-based baseline for tagging part-of-speech and morphosyntactic attributes.
It is competitive with (and complementary to) a supervised character-based model in low-resource settings.
\end{abstract}

\section{Introduction}
\label{sec:intro}

One of the key advantages of word embeddings for natural language processing is that they enable generalization to words that are unseen in labeled training data, by embedding lexical features from large unlabeled datasets into a relatively low-dimensional Euclidean space. These low-dimensional embeddings are typically trained to capture distributional similarity, so that information can be shared among words that tend to appear in similar contexts.

However, it is not possible to enumerate the entire vocabulary of any language, and even large unlabeled datasets will miss terms that appear in later applications. The issue of how to handle these \emph{out-of-vocabulary} (OOV) words poses challenges for embedding-based methods. These challenges are particularly acute when working with low-resource languages, where even unlabeled data may be difficult to obtain at scale. A typical solution is to abandon hope,
by assigning a single OOV embedding to all terms that do not appear in the unlabeled data.



We approach this challenge from a quasi-generative perspective. Knowing nothing of a word except for its embedding and its written form, we attempt to learn the former from the latter. We train a recurrent neural network (RNN) on the character level with the embedding as the target, and use it later to predict vectors for OOV words in any downstream task. We call this model the \ouralgo{}-\rnn, for its ability to read a word's spelling and mimick its distributional embedding.

Through nearest-neighbor analysis, we show that vectors learned via this method capture both word-shape features and lexical features. As a result, we obtain reasonable near-neighbors for OOV abbreviations, names, novel compounds, and orthographic errors. Quantitative evaluation on the Stanford RareWord dataset~\cite{luong2013better} provides more evidence that these character-based embeddings capture word similarity for rare and unseen words.

As an extrinsic evaluation, we conduct experiments on joint prediction of part-of-speech tags and \morsyn\ attributes for a diverse set of 23 languages, as provided in the Universal Dependencies dataset~\cite{de2014universal}. Our model shows significant improvement across the board against a single \textit{UNK}-embedding backoff method, and obtains competitive results against a supervised character-embedding model, which is trained end-to-end on the target task. In low-resource settings, our approach is particularly effective, and is complementary to supervised character embeddings trained from labeled data. The \ouralgo{}-\rnn{} therefore provides a useful new tool for tagging tasks in settings where there is limited labeled data. Models and code are available at \url{www.github.com/yuvalpinter/mimick} .



\section{Related Work}
\label{sec:previous}


\paragraph{Compositional models for embedding rare and unseen words.}
Several studies make use of morphological or orthographic information when training word embeddings, enabling the prediction of embeddings for unseen words based on their internal structure. 
\newcite{botha2014compositional} compute word embeddings by summing over embeddings of the morphemes; \newcite{luong2013better} construct a recursive neural network over each word's morphological parse; \newcite{bhatia2016morphological} use morpheme embeddings as a prior distribution over probabilistic word embeddings. While morphology-based approaches make use of meaningful linguistic substructures, they struggle with names and foreign language words, which include out-of-vocabulary morphemes. Character-based approaches avoid these problems: for example, \newcite{kim2016character} train a recurrent neural network over words, whose embeddings are constructed by convolution over character embeddings; \newcite{wieting2016charagram} learn embeddings of character n-grams, and then sum them into word embeddings. In all of these cases, the model for composing embeddings of subword units into word embeddings is learned by optimizing an objective over a large unlabeled corpus. In contrast, our approach is a post-processing step that can be applied to any set of word embeddings, regardless of how they were trained. This is similar to the ``retrofitting'' approach of \newcite{faruqui2015retrofitting}, but rather than smoothing embeddings over a graph, we learn a function to build embeddings compositionally.





\paragraph{Supervised subword models.}
Another class of methods learn task-specific character-based word embeddings within end-to-end supervised systems. For example, \newcite{santos2014learning} build word embeddings by convolution over characters, and then perform part-of-speech (POS) tagging using a local classifier; the tagging objective drives the entire learning process. \newcite{ling2015finding} propose a multi-level long short-term memory~\cite[LSTM;][]{hochreiter1997long}, in which word embeddings are built compositionally from an LSTM over characters, and then tagging is performed by an LSTM over words. \newcite{plank2016multiling} show that concatenating a character-level or bit-level LSTM network to a word representation helps immensely in POS tagging. Because these methods learn from labeled data, they can cover only as much of the lexicon as appears in their labeled training sets. As we show, they struggle in several settings: low-resource languages, where labeled training data is scarce; morphologically rich languages, where the number of morphemes is large, or where the mapping from form to meaning is complex; and in Chinese, where the number of characters is orders of magnitude larger than in non-logographic scripts. Furthermore, supervised subword models can be combined with \Ouralgo{}, offering additive improvements.


\paragraph{Morphosyntactic attribute tagging.}
We evaluate our method on the task of tagging word tokens for their \morsyn\ attributes, such as gender, number, case, and tense. 
The task of morpho-syntactic tagging dates back at least to the mid 1990s~\cite{oflazer1994tagging,hajivc1998tagging}, and interest has been rejuvenated by the availability of large-scale multilingual morphosyntactic annotations through the Universal Dependencies (UD) corpus~\cite{de2014universal}. For example, \newcite{faruqui2016morpho} propose a graph-based technique for propagating type-level morphological information across a lexicon, improving token-level morphosyntactic tagging in 11 languages, using an SVM tagger. In contrast, we apply a neural sequence labeling approach, inspired by the POS tagger of \newcite{plank2016multiling}.

\section{\Ouralgo{} Word Embeddings}
\label{sec:model}


We approach the problem of out-of-vocabulary (OOV) embeddings as a \textbf{generation} problem: regardless of how the original embeddings were created, we assume there is a generative wordform-based protocol for creating these embeddings. By training a model over the existing vocabulary, we can later use that model for predicting the embedding of an unseen word.

Formally: given a language $\mathcal{L}$, a vocabulary $\mathcal{V} \subseteq \mathcal{L}$ of size $V$, and a pre-trained embeddings table $\mathcal{W} \in \mathbb{R}^{V \times d}$ where each word $\{w_k\}_{k=1}^{V}$ is assigned a vector $\ve_k$ of dimension $d$, our model is trained to find the function $f:\mathcal{L} \rightarrow \mathbb{R}^d$ such that the projected function $f|_{\mathcal{V}}$ approximates the assignments $f(w_k)\approx \ve_k$. Given such a model, a new word $w_{k^*} \in \mathcal{L} \setminus \mathcal{V}$ 
can now be assigned an embedding $\ve_{k^*}=f(w_{k^*})$.

\begin{figure}
\centering
\fbox{\includegraphics[width=6.5cm]{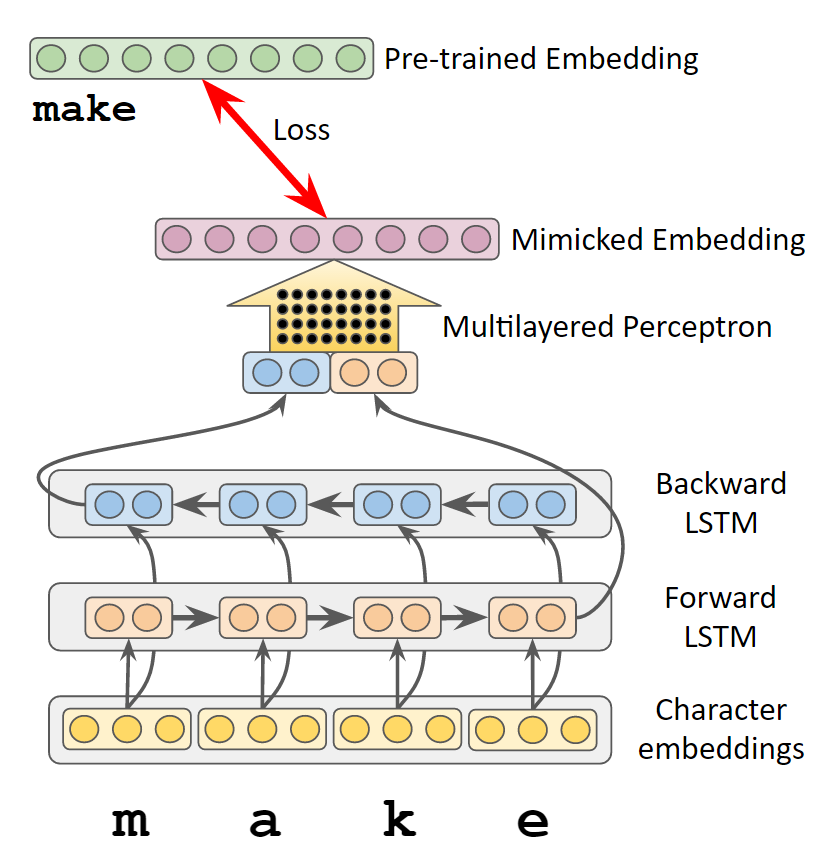}}
\caption{\label{fig:model}\Ouralgo{} model architecture.}
\end{figure}

Our predictive function of choice is a \textbf{Word Type Character Bi-LSTM}. Given a word with character sequence $w=\{c_i\}_1^{n}$, a forward-LSTM and a backward-LSTM are run over the corresponding character embeddings sequence $\{\ve^{(c)}_i\}_1^{n}$.
Let $\vh^n_f$ represent the final hidden vector for the forward-LSTM, and let $\vh^0_b$ represent the final hidden vector for the backward-LSTM. The word embedding is computed by a multilayer perceptron:
\begin{dmath}
f(w)=\mathbf{O}_{T}\cdot g(\mathbf{T}_h\cdot[\vh_{f}^n;\vh_{b}^0]+\vb_h)+\vb_{T},
\label{eq:matching}
\end{dmath}
where $\mathbf{T}_h, \vb_h$ and $\mathbf{O}_{T}, \vb_{T}$ are parameters of affine transformations, and $g$ is a nonlinear elementwise function.
The model is presented in Figure~\ref{fig:model}. 

The training objective is similar to that of \citet{yin2016learning}. We match the predicted embeddings $f(w_k)$ to the pre-trained word embeddings $e_{w_k}$, 
by minimizing the squared Euclidean distance,
\begin{dmath}
\mathcal{L}=\left\lVert f(w_k) - \ve_{w_k} \right\lVert_{2}^{2}.
\label{eq:loss}
\end{dmath}
By backpropagating from this loss, it is possible to obtain local gradients with respect to the parameters of the LSTMs, the character embeddings, and the output model. The ultimate output of the training phase is the character embeddings matrix $\mathbf{C}$ and the parameters of the neural network:
$\mathcal{M}=\{\mathbf{C}, \mathbf{F}, \mathbf{B}, \mathbf{T}_{h}, \vb_{h}, \mathbf{O}_{T}, \vb_{T}\}$, where $\mathbf{F}, \mathbf{B}$ are the forward and backward LSTM component parameters, respectively.

\subsection{\Ouralgo{} Polyglot Embeddings}
\label{ssec:pg-embs}

The pretrained embeddings we use in our experiments are obtained from Polyglot~\cite{al2013polyglot}, a multilingual word embedding effort.
Available for dozens of languages, each dataset contains 64-dimension embeddings for the 100,000 most frequent words in a language's training corpus (of variable size), as well as an \textit{UNK} embedding to be used for OOV words. Even with this vocabulary size, querying words from respective UD corpora (train + dev + test) yields high OOV rates:
in at least half of the 23 languages in our experiments (see Section~\ref{sec:settings}), 29.1\% or more of the word types do not appear in the Polyglot vocabulary. 
The token-level median rate is 9.2\%.%
\footnote{Some OOV counts, and resulting model performance, may be adversely affected by tokenization differences between Polyglot and UD. Notably, some languages such as Spanish, Hebrew and Italian exhibit \textbf{relational synthesis} wherein words of separate grammatical phrases are joined into one form (e.g. Spanish \textit{del = de + el}, `from the-masc.-sg.'). For these languages, the UD annotations adhere to the sub-token level, while Polyglot does not perform sub-tokenization. As this is a real-world difficulty facing users of out-of-the-box embeddings, we do not patch it over in our implementations or evaluation.}

\begin{table*}
\small
\centering
\begin{tabular}{llll}\toprule
\textbf{OOV word} & \textbf{Nearest neighbors} & \textbf{OOV word} & \textbf{Nearest neighbors} \\ 
\midrule
MCT & AWS OTA APT PDM SMP & compartmentalize & formalize rationalize discern prioritize validate \\
McNeally & Howlett Gaughan McCallum Blaney 
& pesky & euphoric disagreeable horrid ghastly horrifying \\
Vercellotti & Martinelli Marini Sabatini Antonelli 
& lawnmower & tradesman bookmaker postman hairdresser \\ 
Secretive & Routine Niche Turnaround Themed 
& developiong & compromising inflating shrinking straining \\ 
corssing & slicing swaying pounding grasping 
& hurtling & splashing pounding swaying slicing rubbing \\
flatfish & slimy jerky watery glassy wrinkle & expectedly & legitimately profoundly strangely energetically \\ 
\bottomrule
\end{tabular}
\caption{Nearest-neighbor examples for the English \ouralgo{} model.}
\label{tab:oov-nns}
\end{table*}

Applying our \ouralgo{} algorithm to Polyglot embeddings, we obtain a prediction model for each of the 23 languages. Based on preliminary testing on randomly selected held-out development sets of 1\% from each Polyglot vocabulary (with error calculated as in Equation~\ref{eq:loss}), we set the following hyper-parameters for the remainder of the experiments: character embedding dimension = 20; one LSTM layer with 50 hidden units; 60 training epochs with no dropout; nonlinearity function $g = \tanh$.\footnote{Other settings, described below, were tuned on the supervised downstream tasks.} We initialize character embeddings randomly, and use DyNet to implement the model~\cite{neubig2017dynet}.


\paragraph{Nearest-neighbor examination.}
As a preliminary sanity check for the validity of our protocol, we examined nearest-neighbor samples in languages for which speakers were available: English, Hebrew, Tamil, and Spanish. Table~\ref{tab:oov-nns} presents selected English OOV words with their nearest in-vocabulary Polyglot words computed by cosine similarity. These examples demonstrate several properties: (a) word shape is learned well (acronyms, capitalizations, suffixes); (b) the model shows robustness to typos (e.g., \textit{develop\textbf{iong}}, \textit{c\textbf{or}ssing}); (c) part-of-speech is learned across multiple suffixes (\textit{pesky -- euphoric, ghastly}); (d) word compounding is detected (e.g., \textit{lawnmower -- bookmaker, postman}); (e) semantics are not learned well (as is to be expected from the lack of context in training), but there are surprises (e.g., \textit{flatfish -- slimy, watery}).
Table~\ref{tab:oov-nns-he} presents examples from Hebrew that show learned properties can be extended to nominal \morsyn\
 attributes (gender, number -- first two examples) and even relational syntactic subword forms such as genetive markers (third example). Names are learned (fourth example) despite the lack of casing in the script. Spanish examples exhibit word-shape and part-of-speech learning patterns with some loose semantics: for example, the plural adjective form \textit{prenatales} is similar to other family-related plural adjectives such as \textit{patrimoniales} and \textit{generacionales}. Tamil displays some semantic similarities as well: e.g. \textit{enjineer} (`engineer') predicts similarity to other professional terms such as \textit{kalviyiyal} (`education'), \textit{thozhilnutpa} (`technical'), and \textit{iraanuva} (`military').

\begin{table*}
\small
\centering
\begin{tabular}{ll}\toprule
\textbf{OOV word} & \textbf{Nearest neighbors} \\ \midrule
TTGFM `(s/y) will come true', & TPTVR `(s/y) will solve', TBTL `(s/y) will cancel', TSIR `(s/y) will remove' \\
GIAVMTRIIM `geometric(m-pl)'$_2$ & ANTVMIIM `anatomic(m-pl)', GAVMTRIIM `geometric(m-pl)'$_1$ \\ 
BQFTNV `our request' & IVFBIHM `their(m) residents', XTAIHM `their(m) sins', IRVFTV `his inheritance' \\
RIC'RDSVN `Richardson' & AVISTRK `Eustrach', QMINQA `Kaminka', GVLDNBRG `Goldenberg' \\
\bottomrule
\end{tabular}
\caption{Nearest-neighbor examples for Hebrew 
(Transcriptions per \newcite{sima2001building}).
`s/y' stands for `she/you-m.sg.'; subscripts denote alternative spellings, standard form being `X'$_1$.
}
\label{tab:oov-nns-he}
\end{table*}

\paragraph{Stanford RareWords.}
\begin{table}
\small
\centering
\begin{tabular}{lcccc}\toprule
 & Emb. & Vocab & Polyglot & All \\
 & dim & size & in-vocab & pairs \\
 & & & $N=862$ & $N=2034$ \\
\midrule
\varembed{} & 128 & 100K & 41.9 & 25.5 \\
Polyglot & 64 & 100K & 40.8 & 8.7 \\
\Ouralgo{} & 64 & 0 & 17.9 & 17.5 \\
Polyglot & \multirow{2}{*}{64} & \multirow{2}{*}{100K} & \multirow{2}{*}{40.8} & \multirow{2}{*}{27.0} \\
~~+\Ouralgo{} \\
Fasttext & 300 & 2.51M & & 47.3 \\
\bottomrule
\end{tabular}
\caption{Similarity results on the RareWord set, measured as Spearman's $\rho \times 100$. \varembed{} was trained on a 20-million token dataset, Polyglot on a 1.7B-token dataset.}
\label{tab:rw}
\end{table}

The Stanford RareWord evaluation corpus~\cite{luong2013better} focuses on predicting word similarity between pairs involving low-frequency English words, predominantly ones with common morphological affixes. As these words are unlikely to be above the cutoff threshold for standard word embedding models, they emphasize the performance on OOV words.

For evaluation of our \ouralgo{} model on the RareWord corpus, we trained the Variational Embeddings algorithm~\cite[VarEmbed;][]{bhatia2016morphological} on a 20-million-token, 100,000-type Wikipedia corpus, obtaining 128-dimension word embeddings for all words in the test corpus. 
\varembed{} estimates a prior distribution over word embeddings, conditional on the morphological composition. For in-vocabulary words, a posterior is estimated from unlabeled data; for out-of-vocabulary words, the expected embedding can be obtained from the prior alone. In addition, we compare to FastText~\cite{fasttext}, a high-vocabulary, high-dimensionality embedding benchmark.

The results, shown in Table~\ref{tab:rw}, demonstrate that the \ouralgo{} \rnn{} recovers about half of the loss in performance incurred by the original Polyglot training model due to out-of-vocabulary words in the ``All pairs'' condition. 
\ouralgo{} also outperforms \varembed{}. FastText can be considered an upper bound: with a vocabulary that is 25 times larger than the other models, it was missing words from only 44 pairs on this data.

\section{Joint Tagging of Parts-of-Speech and \Morsyn\ Attributes}
\label{sec:tag-model}

The Universal Dependencies (UD) scheme~\cite{de2014universal} features a minimal set of 17 POS tags~\cite{petrov2012universal} and supports tagging further language-specific features using attribute-specific inventories. For example, a verb in Turkish could be assigned a value for the evidentiality attribute, one which is absent from Danish. These additional \morsyn\ attributes
are marked in the UD dataset as optional per-token attribute-value pairs.

Our approach for tagging \morsyn\ attributes is similar to the part-of-speech tagging model of \newcite{ling2015finding}, who attach a projection layer to the output of a sentence-level bidirectional LSTM. We extend this approach to \morsyn\ tagging by duplicating this projection layer for each attribute type.
The input to our multilayer perceptron (MLP) projection network is the hidden state produced for each token in the sentence by an underlying LSTM, and the output is attribute-specific probability distributions over the possible values for each attribute on each token in the sequence.
Formally, for a given attribute $a$ with possible values $v \in V_{a}$, the tagging probability for the $i$'th word in a sentence is given by:
\begin{equation}
\Pr(a_{w_i}=v) = (\text{Softmax}(\phi(\vh_i)))_v~,
\end{equation}
with 
\begin{dmath}
\centering
\label{eq:mlp}
\phi(\vh_i) 
= \mathbf{O}_{W}^a\cdot \tanh(\mathbf{W}_h^a\cdot\vh_i+\vb_h^a)+\vb_{W}^a,
\end{dmath}
where $\vh_i$ is the $i$'th hidden state in the underlying LSTM, and $\phi(\vh_i)$ is a two-layer feedforward neural network, with weights $\mathbf{W}^a_h$ and $\mathbf{O}^a_W.$ We apply a softmax transformation to the output; the value at position $v$ is then equal to the probability of attribute $v$ applying to token $w_i$. The input to the underlying LSTM is a sequence of word embeddings, which are initialized to the Polyglot vectors when possible, and to \ouralgo{} vectors when necessary. Alternative initializations are considered in the evaluation, as described in Section~\ref{ssec:models}.


Each tagged attribute sequence (including POS tags) produces a loss equal to the sum of negative log probabilities of the true tags. One way to combine these losses is to simply compute the \textbf{sum loss}.
However, many languages have large differences in sparsity across morpho-syntactic attributes, as apparent from Table~\ref{tab:langs} (rightmost column). We therefore also compute a \textbf{weighted sum loss}, in which each attribute is weighted by the proportion of training corpus tokens on which it is assigned a non-\none{} value.
Preliminary experiments on development set data were inconclusive across languages and training set sizes, and so we kept the simpler sum loss objective for the remainder of our study.
In all cases, part-of-speech tagging was less accurate when learned jointly with \morsyn\ attributes. This may be because the attribute loss acts as POS-unrelated ``noise'' affecting the common LSTM layer and the word embeddings.


\section{Experimental Settings}
\label{sec:settings}


The morphological complexity and compositionality of words varies greatly across languages. While a morphologically-rich agglutinative language such as Hungarian contains words that carry many attributes as fully separable morphemes, a sentence in an analytic language such as Vietnamese may have not a single polymorphemic or inflected word in it.
To see whether this property is influential on our \ouralgo{} model and its performance in the downstream tagging task, we select languages that comprise a sample of multiple morphological patterns. Language family and script type are other potentially influential factors in an orthography-based approach such as ours, and so we vary along these parameters as well. We also considered language selection recommendations from \citet{de2016ud} and \citet{schluter2017empirically}.

\begin{table*}
\small
\centering
\begin{tabular}{p{0.1cm}p{1.2cm}p{0.9cm}p{1.3cm}p{1.0cm}p{1.0cm}p{0.1cm}p{1.0cm}p{1.0cm}p{1.3cm}p{1.0cm}l}
\toprule
 & Language & Branch & Script & Morpho. & Tokens & & Language & Branch & Script & Morpho. & Tokens \\
 &  &  & type &  & w/ attr.  &  & & & type &  & w/ attr. \\
\midrule
vi & Vietnamese & Vietic & alphabetic* & Analytic & 00.0\% & fa & Persian & Iranian & consonantal & Agglutin. & 65.4\% \\
hu & Hungarian & Finno-Ugric & alphabetic & Agglutin. & 83.6\% & hi & Hindi & Indo-Aryan & alphasyllab. & Fusional & 92.4\% \\
id & Indonesian & Malayic & alphabetic & Agglutin. & ---  & lv & Latvian & Baltic & alphabetic & Fusional & 69.2\% \\
zh & Chinese & Sinitic & ideographic & Isolating & 06.2\% & el & Greek & Hellenic & alphabetic & Fusional & 64.8\% \\
tr & Turkish & Turkic & alphabetic & Agglutin. & 68.4\%  & bg & Bulgarian & Slavic & alphabetic & Fusional & 68.6\%  \\
kk & Kazakh & Turkic & alphabetic & Agglutin. & 20.9\%  & ru & Russian & Slavic & alphabetic & Fusional & 69.2\% \\
ar & Arabic & Semitic & consonantal & Fusional & 60.6\%  & cs & Czech & Slavic & alphabetic & Fusional & 83.2\% \\
he & Hebrew & Semitic & consonantal & Fusional & 62.9\%  & es & Spanish & Romance & alphabetic & Fusional & 67.1\% \\
eu & Basque & Vasconic & alphabetic & Agglutin. & 59.2\%  & it & Italian & Romance & alphabetic & Fusional & 67.3\% \\
ta & Tamil & Tamil & syllabic & Agglutin. & 78.8\%  & ro & Romanian & Romance & alphabetic & Fusional & 87.1\% \\
& & & & & & da & Danish & Germanic & alphabetic & Fusional & 72.2\% \\
& & & & & & en & English & Germanic & alphabetic & Analytic & 72.8\% \\
& & & & & & sv & Swedish & Germanic & alphabetic & Analytic & 73.4\% \\
\bottomrule
\end{tabular}
\caption{Languages used in tagging evaluation. Languages on the right are Indo-European.
*In Vietnamese script, whitespace separates syllables rather than words.}
\label{tab:langs}
\end{table*}

As stated above, our approach is built on the Polyglot word embeddings. The intersection of the Polyglot embeddings and the UD dataset (version 1.4) yields 44 languages. Of these, many are under-annotated for \morsyn\ attributes; we select twenty-three  sufficiently-tagged languages, with the exception of Indonesian.\footnote{Vietnamese has no attributes by design; it is a pure analytic language.}
Table~\ref{tab:langs} presents the selected languages
and their typological properties. As an additional proxy for morphological expressiveness, the rightmost column shows
the proportion of UD tokens
which are annotated with any \morsyn\ attribute.

\subsection{Metrics}
\label{ssec:metrics}

As noted above, we use the UD datasets for testing our \ouralgo{} algorithm on 23 languages\footnote{When several datasets are available for a language, we use the unmarked corpus.} with the supplied train/dev/test division. We measure part-of-speech tagging by overall token-level accuracy.

For \morsyn\ attributes, there does not seem to be an agreed-upon metric for reporting performance.
\newcite{dzeroski2000morphosyntactic} report per-tag accuracies on a morphosyntactically tagged  corpus of Slovene.
\newcite{faruqui2016morpho}
report macro-averages of F1 scores of 11 languages from UD 1.1 for the various attributes
(e.g., part-of-speech, case, gender, tense);
recall and precision were calculated for the full set of each attribute's values, pooled together.\footnote{Details were clarified in personal communication with the authors.}
\newcite{agic2013lemmatization}
report separately on parts-of-speech and \morsyn\ attribute accuracies in Serbian and Croatian, as well as precision, recall, and F1 scores per tag.
\newcite{georgiev2012feature}
report token-level accuracy for exact all-attribute tags (e.g. `Ncmsh' for ``Noun short masculine singular definite'') in Bulgarian, reaching a tagset of size 680. 
\newcite{muller2013efficient} do the same for six other languages. We report \textbf{micro F1}
: each token's value for each attribute is compared separately with the gold labeling, where a correct prediction is a matching non-\none{} attribute/value assignment. Recall and precision are calculated over the entire set, with F1 defined as their harmonic mean.

\subsection{Models}
\label{ssec:models}

We implement and test the following models:

\paragraph{No-Char.} Word embeddings are initialized from Polyglot models, with unseen words assigned the Polyglot-supplied \textit{UNK} vector. Following tuning experiments on all languages with cased script, we found it beneficial to first back off to the lowercased form for an OOV word if its embedding exists, and only otherwise assign \textit{UNK}.

\paragraph{\Ouralgo{}.} Word embeddings are initialized from Polyglot, with OOV embeddings inferred from a \ouralgo{} model (Section~\ref{sec:model}) trained on the Polyglot embeddings. Unlike the No-Char case, backing off to lowercased embeddings before using the \ouralgo{} output did not yield conclusive benefits and thus we report results for the more straightforward no-backoff implementation.

\paragraph{\chartotag{}.} Word embeddings are initialized from Polyglot as in the No-Char model (with lowercase backoff), and appended with the output of a character-level LSTM updated during training \cite{plank2016multiling}. This additional module causes a threefold increase in training time.

\paragraph{Both.} Word embeddings are initialized as in \Ouralgo{}, and appended with the \chartotag{} LSTM.

\paragraph{Other models.} Several non-Polyglot embedding models were examined, all performed substantially worse than Polyglot. Two of these are notable: a random-initialization baseline, and a model initialized from FastText embeddings (tested on English). FastText supplies 300-dimension embeddings for 2.51 million lowercase-only forms, and no \textit{UNK} vector.\footnote{Vocabulary type-level coverage for the English UD corpus: 55.6\% case-sensitive, 87.9\% case-insensitive.}
Both of these embedding models were attempted with and without \chartotag{} concatenation.
Another model, initialized from only \ouralgo{} output embeddings, performed well only on the language with smallest Polyglot training corpus (Latvian). A Polyglot model where OOVs were initialized using an averaged embedding of all Polyglot vectors, rather than the supplied \textit{UNK} vector, performed worse than our No-Char baseline on a great majority of the languages.

Last, we do not employ type-based tagset restrictions. All tag inventories are computed from the training sets and each tag selection is performed over the full set.


\begin{table*}
  \small
  \centering
  \begin{tabular}{lllllrlllll}
    \toprule
    & \multicolumn{4}{c}{$N_{train}=5000$} & \multicolumn{4}{c}{Full data} \\
     \cmidrule(lr){2-5} 
    \cmidrule(lr){6-11} 
    
    & No-Char & \Ouralgo & \textsc{char} & Both 
                           & $N_{\text{train}}$ & No-Char & \Ouralgo & \textsc{char} & Both & PSG \\
    & & & \textsc{$\to$tag} & & & & & \textsc{$\to$tag} & & 2016* \\
    \midrule
    kk & --- & --- & --- & --- & 4,949 & 81.94 & 83.95 & 83.64 & 84.88 \\
ta & 82.30 & 81.55 & 84.97 & 85.22 & 6,329 & 80.44 & \textbf{82.96} & 84.11 & 84.46 \\
lv & 80.44 & \textbf{84.32} & 84.49 & \textbf{85.91} & 13,781 & 85.77 & \textbf{87.95} & 89.55 & 89.99 \\
vi & 85.67 & \textit{84.22} & 84.85 & 85.43 & 31,800 & 89.94 & 90.34 & 90.50 & 90.19 \\
hu & 82.88 & \textbf{88.93} & 85.83 & \textbf{88.34} & 33,017 & 91.52 & \textbf{93.88} & 94.07 & 93.74 \\
tr & 83.69 & \textbf{85.60} & 84.23 & \textbf{86.25} & 41,748 & 90.19 & \textbf{91.82} & 93.11 & 92.68 \\
el & 93.10 & \textbf{93.63} & 94.05 & \textbf{94.64} & 47,449 & 97.27 & \textbf{98.08} & 98.09 & 98.22 \\
bg & 90.97 & \textbf{93.16} & 93.03 & \textbf{93.52} & 50,000 & 96.63 & \textbf{97.29} & 97.95 & 97.78 & 98.23 \\
sv & 90.87 & \textbf{92.30} & 92.27 & \textbf{93.02} & 66,645 & 95.26 & \textbf{96.27} & 96.69 & 96.87 & 96.60 \\
eu & 82.67 & \textbf{84.44} & 86.01 & \textbf{86.93} & 72,974 & 91.67 & \textbf{93.16} & 94.46 & 94.29 & 95.38 \\
ru & 87.40 & \textbf{89.72} & 88.65 & \textbf{90.91} & 79,772 & 92.59 & \textbf{95.21} & 95.98 & 95.84 \\
da & 89.46 & 90.13 & 89.96 & 90.55 & 88,980 & 94.14 & \textbf{95.04} & 96.13 & 96.02 & 96.16\\
id & 89.07 & 89.34 & 89.81 & 90.21 & 97,531 & 92.92 & 93.24 & 93.41 & \textbf{93.70} & 93.32 \\
zh & 80.84 & \textbf{85.69} & 81.84 & \textbf{85.53} & 98,608 & 90.91 & \textbf{93.31} & 93.36 & 93.72 \\
fa & 93.50 & 93.58 & 93.53 & 93.71 & 121,064 & 96.77 & \textbf{97.03} & 97.20 & 97.16 & 97.60 \\
he & 90.73 & \textbf{91.69} & 91.93 & 91.70 & 135,496 & 95.65 & \textbf{96.15} & 96.59 & 96.37 & 96.62 \\
ro & 87.73 & \textbf{89.18} & 88.96 & \textbf{89.38} & 163,262 & 95.68 & \textbf{96.72} & 97.07 & 97.09 \\
en & 87.48 & \textbf{88.45} & 88.89 & 88.89 & 204,587 & 93.39 & \textbf{94.04} & 94.90 & 94.70 & 95.17 \\
ar & 89.01 & \textbf{90.58} & 90.49 & 90.62 & 225,853 & 95.51 & \textbf{95.72} & 96.37 & 96.24 & 98.87 \\
hi & 87.89 & 87.77 & 87.92 & 88.09 & 281,057 & 96.31 & 96.45 & 96.64 & 96.61 & 96.97 \\
it & 91.35 & \textbf{92.50} & 92.45 & \textbf{93.01} & 289,440 & 97.22 & 97.47 & 97.76 & 97.69 & 97.90 \\
es & 90.54 & \textbf{91.41} & 91.71 & 91.78 & 382,436 & 94.68 & 94.84 & 95.08 & 95.05 & 95.67 \\
cs & 87.97 & \textbf{90.81} & 90.17 & \textbf{91.29} & 1,173,282 & 96.34 & \textbf{97.62} & 98.18 & \textit{97.93} & 98.02 \\

    \bottomrule
  \end{tabular}
  \caption{POS tagging accuracy (UD 1.4 Test). \textbf{Bold} (\textit{Italic})
indicates significant improvement (degradation)
by McNemar's test, 
$p<.01$, comparing \Ouralgo\ to ``No-Char'', and ``Both'' to \chartotag{}. \\
* For reference, we copy the reported results of \newcite{plank2016multiling}'s analog to \chartotag{}. Note that these were obtained on UD 1.2, and without jointly tagging \morsyn{} \msd{}.}
  \label{tab:pos-acc}
\end{table*}

\begin{table*}
  \small
  \centering
  \begin{tabular}{lllllllll}
    \toprule
    & \multicolumn{4}{c}{$N_{train}=5000$} & \multicolumn{4}{c}{Full data} \\
     \cmidrule(lr){2-5} 
    \cmidrule(lr){6-9}
    & No-Char & \Ouralgo & \textsc{char} & Both & No-Char & \Ouralgo & \textsc{char} & Both \\
    & & & \textsc{$\to$tag} & & & & \textsc{$\to$tag} & \\
    \midrule
    kk & --- & --- & --- & --- & 21.48 & 20.07 & 28.47 & 20.98 \\
ta & 80.68 & \textbf{81.96} & 84.26 & \textbf{85.63} & 79.90 & \textbf{81.93} & 84.55 & 85.01 \\
lv & 56.98 & \textbf{59.86} & 64.81 & \textbf{65.82} & 66.16 & 66.61 & 76.11 & 75.44 \\
hu & 73.13 & \textbf{76.30} & 73.62 & \textbf{76.85} & 80.04 & 80.64 & 86.43 & 84.12 \\
tr & 69.58 & \textbf{75.21} & 75.81 & \textbf{78.93} & 78.31 & \textbf{83.32} & 91.51 & 90.86 \\
el & 86.87 & \textit{86.07} & 86.40 & \textbf{87.50} & 94.64 & \textbf{94.96} & 96.55 & \textbf{96.76} \\
bg & 78.26 & \textbf{81.77} & 82.74 & \textbf{84.93} & 91.98 & \textbf{93.48} & 96.12 & 95.96 \\
sv & 82.09 & \textbf{84.12} & 85.26 & \textbf{88.16} & 92.45 & \textbf{94.20} & 96.37 & \textbf{96.57} \\
eu & 65.29 & \textbf{66.00} & 70.67 & \textit{70.27} & 82.75 & \textbf{84.74} & 90.58 & \textbf{91.39} \\
ru & 77.31 & \textbf{81.84} & 79.83 & \textbf{83.53} & 88.80 & \textbf{91.24} & 93.54 & 93.56 \\
da & 80.26 & \textbf{82.74} & 83.59 & 82.65 & 92.06 & \textbf{94.14} & 96.05 & 95.96 \\
zh & 63.29 & \textbf{71.44} & 63.50 & \textbf{74.66} & 84.95 & 85.70 & 84.86 & 85.87 \\
fa & 84.73 & \textbf{86.07} & 85.94 & 81.75 & 95.30 & \textbf{95.55} & 96.90 & 96.80 \\
he & 75.35 & 68.57 & 81.06 & 75.24 & 90.25 & \textbf{90.99} & 93.35 & 93.63 \\
ro & 84.20 & \textbf{85.64} & 85.61 & \textbf{87.31} & 94.97 & \textbf{96.10} & 97.18 & 97.14 \\
en & 86.71 & \textbf{87.99} & 88.50 & \textbf{89.61} & 95.30 & \textbf{95.59} & 96.40 & 96.30 \\
ar & 84.14 & 84.17 & 81.41 & \textit{81.11} & 94.43 & \textbf{94.85} & 95.50 & 95.37 \\
hi & 83.45 & \textbf{86.89} & 85.64 & 85.27 & 96.15 & 96.21 & 96.59 & \textbf{96.67} \\
it & 89.96 & \textbf{92.07} & 91.27 & \textbf{92.62} & 97.32 & \textbf{97.80} & 98.18 & 98.31 \\
es & 88.11 & \textbf{89.81} & 88.58 & \textbf{89.63} & 94.84 & \textbf{95.44} & 96.21 & \textbf{96.84} \\
cs & 68.66 & \textbf{72.65} & 71.02 & \textbf{73.61} & 91.75 & \textbf{93.71} & 95.29 & 95.31 \\


    \bottomrule
  \end{tabular}
  \caption{Micro-F1 for \morsyn{} \msd{} (UD 1.4 Test). \textbf{Bold} (\textit{Italic}) type indicates significant improvement (degradation) by a bootstrapped $Z$-test, $p<.01$, comparing
  models as in Table~\ref{tab:pos-acc}. Note that the Kazakh (\textit{kk}) test set has only 78 morphologically tagged tokens.}
  \label{tab:att-f1}
\end{table*}

\begin{figure*}
\centering
\begin{tabular}{cc}
\includegraphics[width=7.9cm]{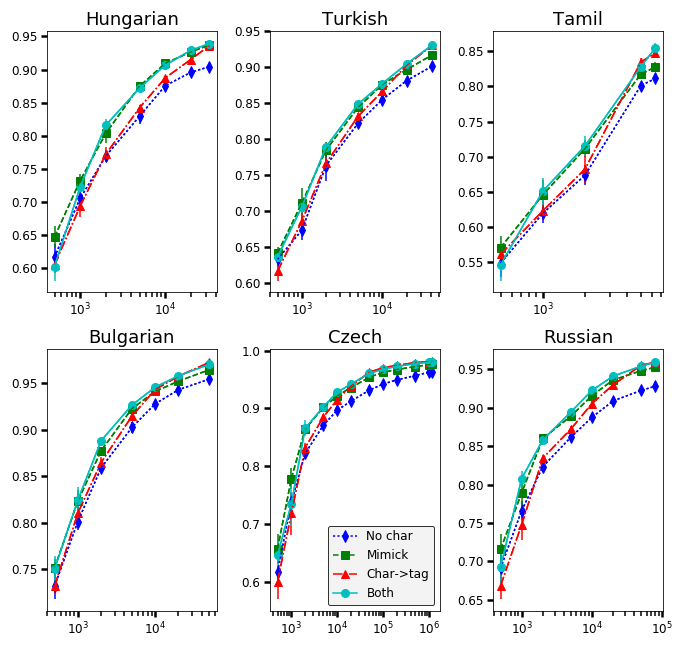} &
\includegraphics[width=7.9cm]{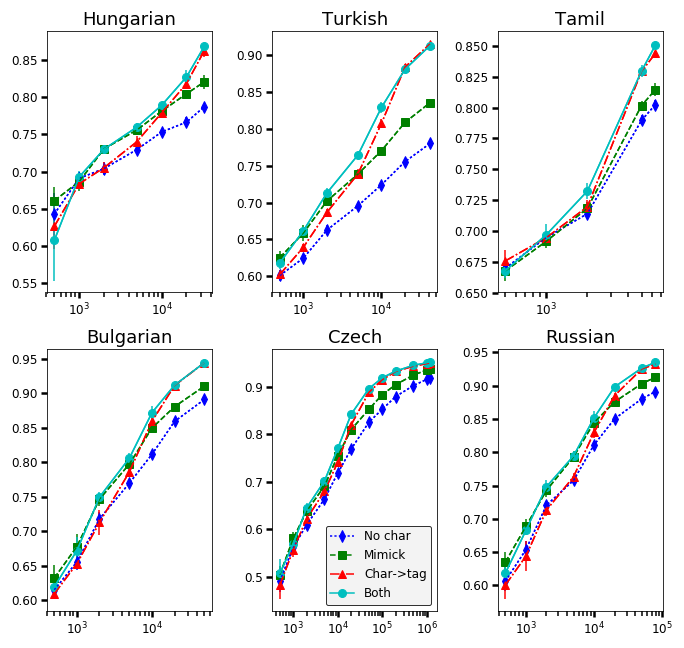} \\
Part-of-speech tagging (accuracy) & Morpho-syntactic attribute tagging (micro-F1)\\
\end{tabular}
\caption{Results on agglutinative languages (top) and on Slavic languages (bottom).
X-axis is number of training tokens, starting at 500. Error bars are the standard deviations over five random training data subsamples.}
\label{fig:agg-slav}
\end{figure*}



\subsection{Hyperparameters}
Based on development set experiments, we set the following hyperparameters for all models on all languages: two LSTM layers of hidden size 128, MLP hidden layers of size equal to the number of each attribute's possible values; momentum stochastic gradient descent with $0.01$ learning rate; 40 training epochs (80 for 5K settings) with a dropout rate of 0.5. The \chartotag{} models use 20-dimension character embeddings and a single hidden layer of size 128. 

\section{Results}

We report performance in both low-resource and full-resource settings. Low-resource training sets were obtained by randomly sampling training sentences, without replacement, until a predefined token limit was reached. We report the results on the full sets and on $N=5000$ tokens in Table~\ref{tab:pos-acc} (part-of-speech tagging accuracy) and Table~\ref{tab:att-f1} (\morsyn\ attribute tagging micro-F1).
Results for additional training set sizes are shown in \autoref{fig:agg-slav}; space constraints prevent us from showing figures for all languages.


\paragraph{\Ouralgo{} as OOV initialization.}
In nearly all experimental settings on both tasks,
across languages and training corpus sizes, the \ouralgo{} embeddings significantly improve over the Polyglot \textit{UNK} embedding for OOV tokens on both POS and \morsyn\ tagging.
For POS, the largest margins are in the Slavic languages (Russian, Czech, Bulgarian), where word order is relatively free and thus rich word representations are imperative. Chinese also exhibits impressive improvement across all settings, perhaps due to the large character inventory ($>$ 12,000), for which a model such as \ouralgo{} can learn well-informed embeddings using the large Polyglot vocabulary dataset, overcoming both word- and character-level sparsity in the UD corpus.\footnote{Character coverage in Chinese Polyglot is surprisingly good: only eight characters from the UD dataset are unseen in Polyglot, across more than 10,000 unseen word types.} In \morsyn\ tagging, gains are apparent for Slavic languages and Chinese, but also for agglutinative languages --- especially Tamil and Turkish --- where the stable morpheme representation makes it easy for subword modeling to provide a type-level signal.\footnote{Persian is officially classified as agglutinative but it is mostly so with respect to derivations. Its word-level inflections are rare and usually fusional.} To examine the effects on Slavic and agglutinative languages in a more fine-grained view, we present results of multiple training-set size experiments for each model, averaged over five repetitions (with different corpus samples), in Figure~\ref{fig:agg-slav}.

\paragraph{\Ouralgo{} vs. \chartotag{}.} In several languages, the \ouralgo{} algorithm fares better than the \chartotag{} model on part-of-speech tagging in low-resource settings.
Table~\ref{tab:10k} presents the POS tagging improvements that \ouralgo{} achieves over the pre-trained Polyglot models, with and without \chartotag{} concatenation, with 10,000 tokens of training data.
We obtain statistically significant improvements in most languages, even when \chartotag{} is included. 
These improvements are particularly substantial for test-set tokens outside the UD training set, as shown in the right two columns. While test set OOVs are a strength of the \chartotag{} model \cite{plank2016multiling}, in many languages there are still considerable improvements to be obtained from the application of \ouralgo{} initialization. This suggests that with limited training data, the end-to-end \chartotag{} model is unable to learn a sufficiently accurate representational mapping from orthography.

\begin{table}
\small
\centering
\small
\begin{tabular}{p{1.3cm}rrrrr}\toprule
Test set & \multicolumn{1}{c}{Missing} & \multicolumn{2}{c}{Full} & \multicolumn{2}{c}{OOV}\\
 \multicolumn{2}{r}{embeddings} & \multicolumn{2}{c}{vocabulary} & \multicolumn{2}{c}{(UD)} \\
\midrule
\chartotag{} & & w/o & with & w/o & with \\ \midrule
Persian & 2.2\% & 0.03 & \textbf{0.41} & \textbf{0.83} & \textbf{0.81} \\
Hindi & 3.8\% & \textbf{0.59} & 0.21 & \textbf{3.61} & 0.36 \\
English & 4.5\% & \textbf{0.83} & 0.25 & \textbf{3.26} & 0.49 \\
Spanish & 5.2\% & 0.33 & -0.26 & 1.03 & -0.66 \\
Italian & 6.6\% & \textbf{0.84} & 0.28 & \textbf{1.83} & 0.21 \\
Danish & 7.8\% & 0.65 & \textbf{0.99} & \textbf{2.41} & \textbf{1.72} \\
Hebrew & 9.2\% & \textbf{1.25} & \textbf{0.40} & \textbf{3.03} & 0.06 \\
Swedish & 9.2\% & \textbf{1.50} & \textbf{0.55} & \textbf{4.75} & \textbf{1.79} \\
Bulgarian & 9.4\% & \textbf{0.96} & 0.12 & \textbf{1.83} & -0.11 \\
Czech & 10.6\% & \textbf{2.24} & \textbf{1.32} & \textbf{5.84} & \textbf{2.20} \\
Latvian & 11.1\% & \textbf{2.87} & \textbf{1.03} & \textbf{7.29} & \textbf{2.71} \\
Hungarian & 11.6\% & \textbf{2.62} & \textbf{2.01} & \textbf{5.76} & \textbf{4.85} \\
Turkish & 14.5\% & \textbf{1.73} & \textbf{1.69} & \textbf{3.58} & \textbf{2.71} \\
Tamil* & 16.2\% & \textbf{2.52} & 0.35 & 2.09 & 1.35 \\  
Russian & 16.5\% & \textbf{2.17} & \textbf{1.82} & \textbf{4.55} & \textbf{3.52} \\
Greek & 17.5\% & \textbf{1.07} & 0.34 & \textbf{3.30} & 1.17 \\
Indonesian & 19.1\% & \textbf{0.46} & 0.25 & \textbf{1.19} & 0.75 \\
Kazakh* & 21.0\% & 2.01 & 1.24 & \textbf{5.34} & \textbf{4.20} \\ 
Vietnamese & 21.9\% & 0.53 & \textbf{1.18} & 1.07 & \textbf{5.73} \\
Romanian & 27.1\% & \textbf{1.49} & \textbf{0.47} & \textbf{4.22} & \textbf{1.24}\\
Arabic & 27.1\% & \textbf{1.23} & \textbf{0.32} & \textbf{2.15} & 0.22 \\
Basque & 35.3\% & \textbf{2.39} & \textbf{1.06} & \textbf{5.42} & \textbf{1.68} \\
Chinese & 69.9\% & \textbf{4.19} & \textbf{2.57} & \textbf{9.52} & \textbf{5.24} \\

\bottomrule
\end{tabular}
\caption{Absolute gain in POS tagging accuracy from using \ouralgo{} for 10,000-token datasets (all tokens for Tamil and Kazakh).
\textbf{Bold} denotes statistical significance (McNemar's test,
$p<0.01$).}
\label{tab:10k}
\end{table}

\section{Conclusion}
\label{sec:conclusion}

We present a straightforward algorithm to infer OOV word embedding vectors from pre-trained, limited-vocabulary models, without need to access the originating corpus.
This method is particularly useful for low-resource languages and tasks with little labeled data available,
and in fact is task-agnostic. Our method
improves performance over word-based models on annotated sequence-tagging tasks for a large variety of languages across dimensions of family, orthography, and morphology.
In addition, we present a Bi-LSTM approach for tagging morphosyntactic attributes at the token level. In this paper, the \ouralgo{} model was trained using characters as input, but future work may consider the use of other subword units, such as morphemes, phonemes, or even bitmap representations of ideographic characters~\cite{costa2017chinese}.
\section{Acknowledgments}
We thank Umashanthi Pavalanathan, Sandeep Soni, Roi Reichart, and our anonymous reviewers for their valuable input.
We thank Manaal Faruqui and Ryan McDonald for their help in understanding the metrics for \morsyn\ tagging. The project was supported by project HDTRA1-15-1-0019 from the Defense Threat Reduction Agency.

\clearpage

\bibliographystyle{emnlp_natbib.bst}
\bibliography{char-matching,cite-strings,cites,cite-definitions}

\end{document}